\pdfoutput=1
\PassOptionsToPackage{draft}{minted}
\documentclass{arxivtmpl}

\let\templatebibliographystyle\bibliographystyle
\renewcommand{\bibliographystyle}[1]{}

\def\1{\bm{1}}

\DeclareMathAlphabet{\mathsfit}{\encodingdefault}{\sfdefault}{m}{sl}
\SetMathAlphabet{\mathsfit}{bold}{\encodingdefault}{\sfdefault}{bx}{n}

\tcbuselibrary{skins,breakable}
\newtcolorbox{findingbox}{enhanced,breakable,frame hidden,colback=gred!6,borderline west={3pt}{0pt}{gred},sharp corners,boxsep=0pt,left=12pt,right=10pt,top=8pt,bottom=8pt,before skip=12pt,after skip=12pt}
\newtcolorbox{notebox}{enhanced,breakable,frame hidden,colback=gblue!6,borderline west={3pt}{0pt}{gblue},sharp corners,boxsep=0pt,left=12pt,right=10pt,top=8pt,bottom=8pt,before skip=12pt,after skip=12pt}
\newtcolorbox{tipbox}{enhanced,breakable,frame hidden,colback=ggreen!6,borderline west={3pt}{0pt}{ggreen},sharp corners,boxsep=0pt,left=12pt,right=10pt,top=8pt,bottom=8pt,before skip=12pt,after skip=12pt}
\newtcolorbox{warnbox}{enhanced,breakable,frame hidden,colback=gyellow!6,borderline west={3pt}{0pt}{gyellow},sharp corners,boxsep=0pt,left=12pt,right=10pt,top=8pt,bottom=8pt,before skip=12pt,after skip=12pt}

\tcbuselibrary{listings}
\newcommand{\boxfont}{\fontsize{9}{11}\selectfont}
\newtcblisting{promptbox}[1][]{%
  enhanced, listing only, breakable, frame hidden, width=\linewidth,
  colback=gblue!6, borderline west={3pt}{0pt}{gblue},
  sharp corners, boxsep=0pt,
  left=12pt, right=10pt, top=8pt, bottom=8pt, before skip=12pt, after skip=12pt,
  before upper={\def\promptlabel{#1}\ifx\promptlabel\empty\else{\sffamily\bfseries\color{gblue}#1}\par\smallskip\fi},   
  listing options={basicstyle=\ttfamily\boxfont, escapeinside={(*}{*)}},
}

\newtcolorbox{transcriptbox}{%
  enhanced, breakable, frame hidden,
  colback=ggreen!6, borderline west={3pt}{0pt}{ggreen},
  sharp corners, boxsep=0pt, fontupper=\ttfamily\boxfont, halign upper=flush left,   
  left=12pt, right=10pt, top=6pt, bottom=8pt, before skip=12pt, after skip=12pt,
}

\colorlet{codebg}{gyellow!6}
\setminted{fontsize=\boxfont, breaklines, autogobble, tabsize=4, style=vs}
\newminted[codebox]{python}{bgcolor=codebg, frame=leftline, framerule=2.5pt, rulecolor=gyellow, framesep=2.5mm}

\hypersetup{
    colorlinks=true,
    linkcolor=linkcol,
    citecolor=linkcol,
    urlcolor=linkcol,
    pdfborderstyle={},
    pdfborder={0 0 0}
}
\renewcommand\Affilfont{\rmfamily\mdseries\fontsize{11.5}{14}\selectfont}

\colorlet{perfhue}{gblue}
\colorlet{perf0}{white}
\colorlet{perf10}{perfhue!12}
\colorlet{perf20}{perfhue!22}
\colorlet{perf30}{perfhue!33}
\colorlet{perf40}{perfhue!44}
\colorlet{perf50}{perfhue!55}
\colorlet{perf60}{perfhue!66}
\colorlet{perf70}{perfhue!77}
\colorlet{perf80}{perfhue!88}

\newcommand{\RN}[1]{%
	\textup{\lowercase\expandafter{\it \romannumeral#1}}%
}

\SetKwInput{KwInput}{Input}
\SetKwInput{KwOutput}{Output}
\DontPrintSemicolon

\SetKwComment{Comment}{\color{ggreen}\# }{}

\SetKwProg{Function}{def}{:}{}

\SetKwProg{For}{for}{:}{}
\SetKwProg{If}{if}{:}{}

\let\bibliographystyle\templatebibliographystyle
\usepackage[T1]{fontenc}
\newcommand*{\codedatanotetext}{}
\newcommand{\codedatanote}[1]{\gdef\codedatanotetext{#1}}
\renewcommand{\footnoterule}{\kern-3pt\hrule width\columnwidth\kern2.6pt}

\titlespacing*{\section}{0pt}{1.1\baselineskip}{0.35\baselineskip}
\titlespacing*{\subsection}{0pt}{0.85\baselineskip}{0.15\baselineskip}
\titleformat{\section}{\rmfamily\bfseries\Large}{\thesection}{0.7em}{}
\titleformat{\subsection}{\rmfamily\bfseries\large}{\thesubsection}{0.7em}{}
\setlist[itemize]{topsep=4pt,itemsep=3pt,parsep=0pt}

\renewcommand{\arraystretch}{1.08}

\fancypagestyle{fancy}{%
  \fancyhf{}%
  \fancyhead[C]{\fontsize{9}{11}\selectfont\color{black!50}%
    SlackBench: Benchmarking Agents on Collaborative Projects Grounded in Real Code Repositories}%
  \fancyfoot[C]{\small\thepage}%
  \renewcommand{\footrulewidth}{0pt}}
\makeatletter
\fancypagestyle{firstpagestyle}{%
  \fancyhf{}%
  \fancyfoot[L]{\codedatanotetext}%
  \renewcommand{\footrulewidth}{\ifx\codedatanotetext\@empty 0pt\else 0.3pt\fi}}
\makeatother

\renewcommand{\titlefont}{\normalfont\bfseries\fontsize{22}{26}\selectfont}

\renewcommand\Affilfont{\rmfamily\mdseries\fontsize{10.5}{13}\selectfont}

\makeatletter
\renewcommand\AB@authnote[1]{\expandafter\ifblank\expandafter{#1}{}{\,\textsuperscript{\normalfont#1}}}
\renewcommand\AB@affilnote[1]{\textsuperscript{\normalfont#1}}
\renewcommand{\maketitle}{%
  \begingroup
  \setlength{\parindent}{0pt}%
  \raggedright
  \vspace*{-12pt}%
  {\titlefont\exhyphenpenalty=0 \@title\par}%
  \vspace{12pt}%
  {\@author\par}%
  \ifx\@authornote\@empty\else
    \vspace{5pt}%
    {\fontsize{9}{11.5}\selectfont\@authornote\par}%
  \fi
  \vspace{12pt}%
  \endgroup}
\makeatother

\renewenvironment{abstract}
  {\par\begingroup\small\noindent\textbf{Abstract}\par\nobreak\vspace{3pt}\noindent\ignorespaces}
  {\par\endgroup\vspace{5pt}}

\hypersetup{
  pdftitle={SlackBench: Benchmarking Agents on Collaborative Projects Grounded in Real Code Repositories},
  pdfauthor={Mingxuan Li, Huanzhi Mao, Ruitao Zheng, Jarod Stanbury James, Yunze Li, Natanya Anderson, Joseph E. Gonzalez, Alex Dimakis},
  pdfsubject={Benchmark for agents on collaborative research and engineering workflows over workplace communication and code repositories},
  pdfkeywords={SlackBench, LLM agents, agentic search, abstention, summarization, Slack, code repositories, benchmark},
  bookmarksnumbered=true}

\title{SlackBench: Benchmarking Agents on\\ Collaborative Projects Grounded in\\ Real Code Repositories}

\author[1,2,3$*$]{Mingxuan~Li}
\author[1$*$]{Huanzhi~Mao}
\author[1]{Ruitao~Zheng}
\author[1]{Jarod~Stanbury~James}
\author[1]{Yunze~Li}
\author[1]{Natanya~Anderson}
\author[1]{Joseph~E.~Gonzalez}
\author[1,3]{Alex~Dimakis}
\affil[1]{\mbox{University of California, Berkeley}}
\affil[2]{\mbox{University of Chicago}}
\affil[3]{\mbox{Bespoke Labs}}
\authornote{$*$\,Equal contribution.\quad Correspondence to: \texttt{huanzhimao@berkeley.edu}.}

\begin{document}

\maketitle
\setcounter{footnote}{1}

\thispagestyle{firstpagestyle}

\begin{abstract}
AI Agents are becoming digital coworkers that can browse the web, write code, run experiments, and assist humans over extended horizons. Yet existing benchmarks fall short of real knowledge work: many rely on synthetic data or isolated tasks, missing the complex, multi-party dynamics of collaborative projects, where critical information is scattered across conversations, code, and evolving decisions. Even simple workplace questions may require locating relevant channels and direct messages, recovering context from long discussions, inspecting repository state, reconciling stale or conflicting evidence, and synthesizing across sources.

We introduce SlackBench, a benchmark for evaluating agents on realistic research and engineering workflows that require joint reasoning over workplace communication and evolving repository state. Grounded in privacy-preserving reenactments of real academic research projects, SlackBench provides reproducible environments with realistic multi-party dialogue, evolving decisions, and corresponding code changes grounded in GitHub repositories and Google Docs. It contains 140 queries spanning heterogeneous-source agentic search, abstention on unanswerable or false-premise questions, goal-directed summarization, and circumstance inference over long conversations. These tasks are not reducible to static corpus question answering: agents must determine which conversations matter, traverse threads, inspect repository state, and synthesize evidence across communication and code artifacts. Across 33 frontier models and agent harnesses, the best system scores around 80\% and most agents score below 66\%, highlighting substantial headroom for improvement. SlackBench thus provides an extensible framework for evaluating agents in collaborative work environments.\codedatanote{\scriptsize The SlackBench dataset is available at \url{https://huggingface.co/datasets/HuanzhiMao/SlackBench}.}
\end{abstract}

\section{Introduction}
\label{sec:intro}

Large language model agents are increasingly positioned as digital coworkers that can write code, run experiments, interact with diverse workplace applications, browse the web, and assist humans over extended horizons. Yet the benchmarks used to measure this progress still capture only fragments of real knowledge work. In practice, project state is scattered across Slack channels, Google Docs, GitHub commits, source files, experiment outputs, and informal decisions that evolve over time. Answering a seemingly simple workplace question often requires deciding where to look, recovering context from long multi-party discussions, reconciling stale or conflicting statements, inspecting the codebase, browsing the web, and synthesizing evidence across all of these sources.

Existing evaluation suites have improved realism for web navigation, software engineering, workplace simulation, and long-horizon memory \citep{wei2025browsecompsimplechallengingbenchmark,jimenez2024swebenchlanguagemodelsresolve}, but they typically isolate only one or two components of this process. On the other hand, benchmarks that study agents in multi-platform professional settings rarely combine heterogeneous workplace applications with conversations grounded in realistic collaborative scenarios, evolving decisions, and concrete project artifacts \citep{xu2025theagentcompanybenchmarkingllmagents,deshpande2025memtrackevaluatinglongtermmemory}. We therefore lack a shareable and reproducible benchmark of grounded collaborative workspaces in which agents must reason across communication, documents, code, and repository history. This gap is especially acute for deploying agent assistants in research and engineering teams, where important context is rarely contained in a single document, conversation, or repository snapshot.

To address this gap, we present SlackBench, a benchmark and evaluation environment built from privacy-preserving, human-in-the-loop simulations of academic research workflows, grounded in concrete project artifacts such as GitHub repositories and Google Docs. This process produces realistic multi-party dialogues tied to evolving code, documents, experiment results, and project decisions, while avoiding the release of private organizational chat histories. The resulting environments preserve chronology, task dependencies, artifact evolution, and conversational realism while remaining shareable and reproducible.

SlackBench contains 140 tasks spanning heterogeneous-source agentic search, abstention from hallucination under unanswerable or false-premise questions, goal-directed summarization, and circumstance inference from long, multi-party dialogues. These tasks go beyond static corpus question answering: agents must identify relevant channels and direct messages, follow threads, inspect code and commit history, resolve stale or conflicting evidence, and synthesize grounded answers across communication, documents, and repository artifacts. In many cases, single-shot retrieval over a flattened corpus is insufficient.

We evaluate 33 frontier models and agent harnesses on SlackBench and find that the best systems achieve around 80\% overall while most agents score below 66\%. 
Our contributions are threefold:
\begin{itemize}
    \item We introduce a benchmark of 140 challenging tasks grounded in reconstructed, artifact-backed project histories inspired by real Slack, GitHub, and Google Docs workflows (\cref{sec:slackbench}).
    
    \item We conduct a systematic study on how different harnesses and ways of accessing the data affect agent performance. We perform more than 40 
    experiments ablating across different models, harnesses, and reasoning efforts. One surprising finding is that directly giving the data as files in the CLI is more efficient and performant compared to MCP access (\cref{para:cli_vs_mcp}).
    
    \item We perform an extensive failure analysis. We analyze agent failure modes across task categories, revealing recurring weaknesses in insufficient global verification, false-premise recognition, holding on to incorrect proxies, etc., that can guide future improvements to workplace agents. 
\end{itemize}
\section{Related Work}
\label{sec:related_work}

\textbf{Agent and tool-use benchmarks:} Early agent benchmarks such as AgentBench \citep{liu2025agentbenchevaluatingllmsagents} established multi-environment interactive evaluation for LLM agents. A parallel line of work focused on tool use: API-Bank \citep{li2023apibankcomprehensivebenchmarktoolaugmented} studies API planning and calling; ToolLLM \citep{qin2023toolllmfacilitatinglargelanguage} scales tool-use data to thousands of APIs; BFCL \citep{patil2025bfcl} standardizes function-calling evaluation; and ToolSandbox \citep{lu2025toolsandboxstatefulconversationalinteractive} introduces stateful, conversational tool execution. $\tau$-bench \citep{yao2024taubenchbenchmarktoolagentuserinteraction} further extends this line by evaluating dynamic user–agent interaction under domain-specific policies. These benchmarks are foundational for measuring action selection and tool correctness, but they generally do not require agents to recover latent project state from long, noisy, multi-party workplace histories dispersed across communication and code.

\textbf{Multi-platform enterprise, memory, and personal-context benchmarks:} Existing work closest to SlackBench include AutomationBench \citep{shepard2026automationbench}, APEX-Agents \citep{vidgen2026apexagents}, and Workspace-Bench \citep{tang2026workspacebench10benchmarkingai}. AutomationBench evaluates agent capabilities to act in relatively small workspace environments, with a median of 3.3k tokens in environment. SlackBench's Slack environment alone has 790k tokens. APEX-Agents evaluate agents in specific domains of banking, consulting, and law. Workspace-Bench uses completely synthetic environment, and often instruct agent to directly look at specific files or to ignore specific files. MEMTRACK \citep{deshpande2025memtrackevaluatinglongtermmemory} interleaves events across workplace tools to evaluate long-term memory and state tracking in multi-platform organizational workflows, while Agent-Diff \citep{pysklo2026agentdiffbenchmarkingllmagents} evaluates enterprise API tasks through code execution and state-diff-based success criteria. Personalized assistant benchmarks such as ASTRA-bench \citep{xiu2026astrabenchevaluatingtooluseagent} study user profiles or time-evolving personal context in tool use. SlackBench complements these directions by focusing on collaborative project understanding: answers depend not only on remembering facts, but on discovering which conversations matter, reconciling social and technical evidence, abstaining when the premise is false, and inferring roles or dispositions from multi-party behavior.

\textbf{Software-engineering benchmarks and harness evaluation:} SlackBench also relates to software-engineering agent evaluation. SWE-bench \citep{jimenez2024swebenchlanguagemodelsresolve} showed that real GitHub issues provide a demanding testbed for autonomous code work, and SWE-agent \citep{yang2024sweagentagentcomputerinterfacesenable} demonstrated that agent-computer interface design can substantially improve performance. OpenHands \citep{wang2025openhandsopenplatformai} provides a flexible platform for building and evaluating software agents, MLE-bench \citep{chan2025mlebenchevaluatingmachinelearning} extends evaluation to machine-learning engineering, and Terminal-Bench \citep{merrill2026terminalbenchbenchmarkingagentshard} targets hard terminal-based tasks inspired by real workflows. More recently, SkillsBench \citep{li2026skillsbenchbenchmarkingagentskills} and AgencyBench \citep{li2026agencybenchbenchmarkingfrontiersautonomous} explicitly study the effect of skills or scaffolds, reinforcing that agent performance depends on more than base-model quality alone. Our paired base-model-versus-harness evaluation follows this line, but in a socio-technical workflow where success depends jointly on conversational navigation and repository interaction.

While recent benchmarks study tool use, software engineering, workplace automation, and multi-platform memory, they do not directly evaluate agents on privacy-preserving reconstructions of real collaborative project histories in which answers must be assembled from multi-party Slack-style communication and GitHub repository.

\begin{figure*}[t]
    \centering
    \includegraphics[width=1\textwidth]{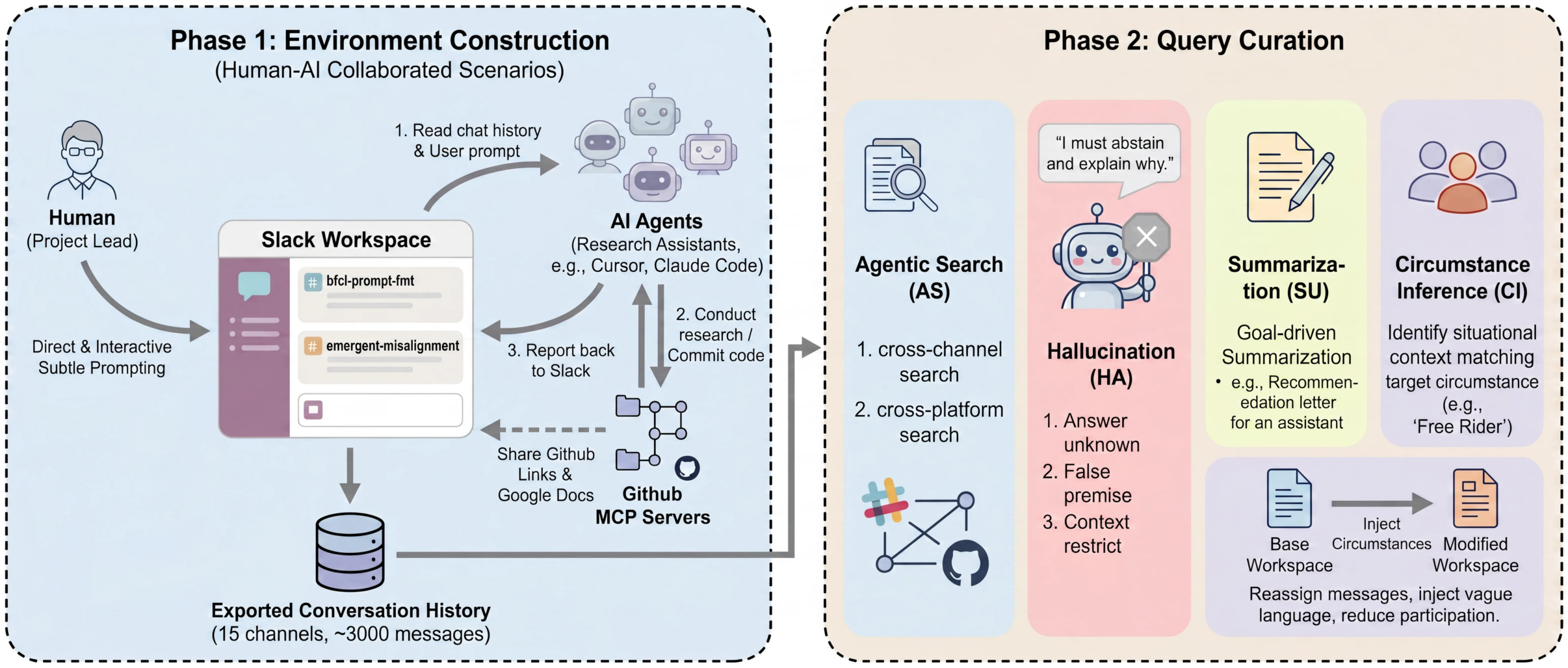}
    \caption{Overview of the SlackBench benchmark. SlackBench evaluates coding agents on realistic Slack workspace tasks spanning agentic search, abstention, summarization, and circumstance inference in complex and heterogeneous scenarios.}
    \label{fig:slackbench}
\end{figure*}

\section{Data Curation}
\label{sec:slackbench}

\subsection{Environment Curation}
\label{subsec:env_curation}

Our benchmark is built on real Slack conversations. We generate the conversations directly in a live Slack workspace and then export the full conversation history to a local directory for filtering and evaluation. To make the conversations realistic, we leverage both human effort and LLM-powered agents to simulate genuine research scenarios. Of all the Slack messages, around 35.0\% were directly authored by humans, and another 35.2\% were substantially rewritten by humans.

Concretely, as shown in \cref{fig:slackbench}, each channel in our Slack workspace corresponds to a real research project initiated from scratch around a concrete research idea (e.g., investigating the sensitivity of LLM tool-call accuracy to prompt format, as part of the Berkeley Function-Calling Leaderboard \citep{patil2025bfcl}). Conversations are produced by having humans and coding agents such as Cursor and Claude Code carry out—and in some cases complete—these research projects, using Slack as the primary communication platform. Humans play the role of project leads, while multiple agent sessions act as research assistants reporting to them. We create a Slack account and a GitHub account for each agent session and connect the agent to Slack and GitHub MCP servers \citep{anthropic2024mcp}.

To ensure that the Slack conversations are grounded in real, concrete project data, we connect each Slack channel to one or more GitHub repositories containing project code and writing. Some channels also include Google Docs to mimic real meeting notes, IRB applications, and similar artifacts. This integration is achieved by having humans or agents share URLs of GitHub repositories and Google Docs in Slack in natural ways that resemble real research conversations.

Additionally, to keep the simulation close to real-world scenarios, we use both direct system prompting and subtle in-context prompting to control the non-coding behavior of the agents, including Slack messaging and GitHub operations. With direct system prompting, we instruct the model to communicate like a human research assistant and to follow a specified workflow when prompted. However, such workflows can be rigid and fail to capture the diverse conversational dynamics of a real research group. To mitigate this, we additionally embed subtle hints in the user prompts within the agent's interactive interface to guide the conversation flow more naturally.

After generation finishes for the whole workspace, we export the entire workspace chat history and perform manual filtering of the Slack messages: rewriting messages that sound too AI-like, replacing real API keys with fake ones, correcting hallucinated content against the GitHub repositories, and similar cleanup.

\begin{figure*}[t]
    \centering
    \includegraphics[width=1\textwidth]{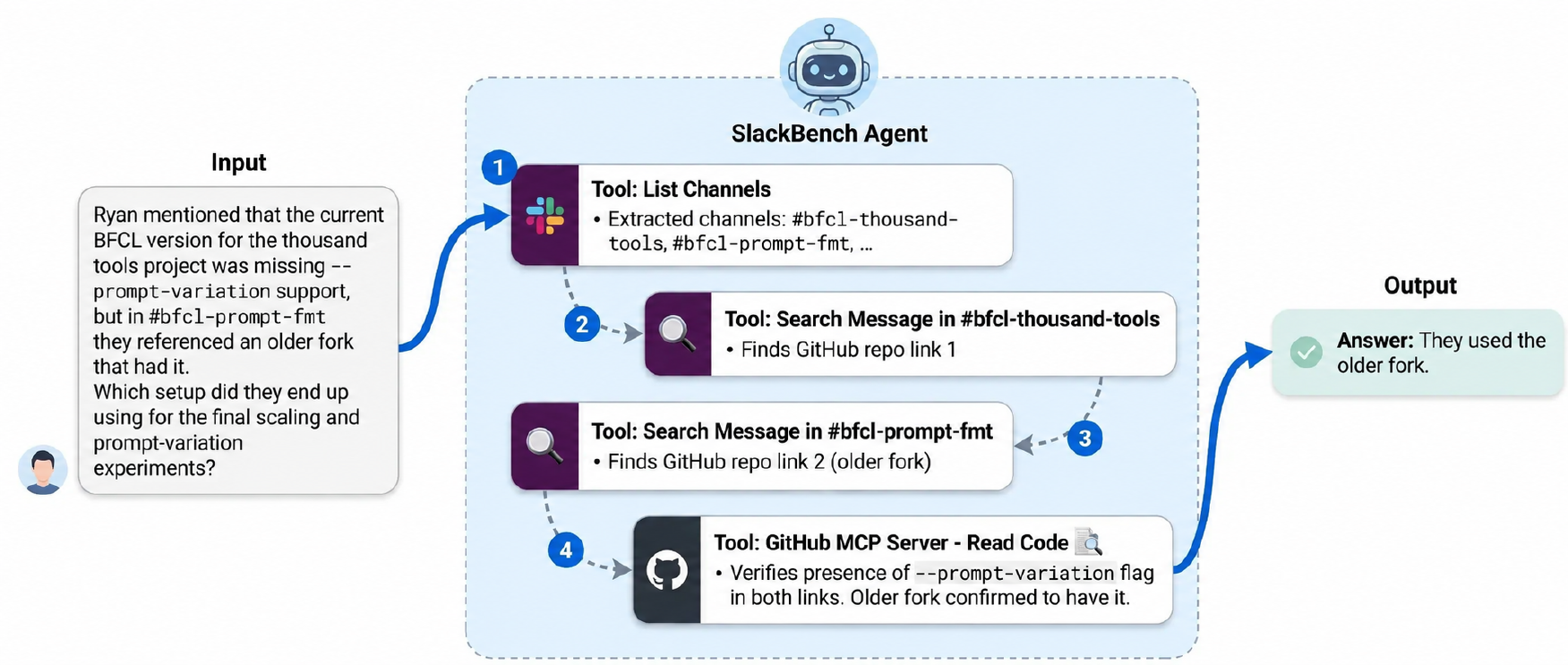}
    \caption{An example agent trajectory on a SlackBench query, illustrating the sequence of tool calls and reasoning steps an agent takes to retrieve information from the Slack workspace.}
    \label{fig:trajectory_show}
\end{figure*}

We generate a total of 15 channels with around 3000 messages. We also have 14 real GitHub repositories and 9 Google Docs with links in the Slack conversations. We include summaries of each channel and more details in \cref{appendix:chat_gen}. 

\subsection{Query Curation}
\label{subsec:query_curation}

To systematically evaluate the performance of LLMs and agent harnesses on realistic Slack-based queries, we construct four categories of queries targeting different capabilities. Example queries for each category are shown in \cref{tab:example_queries_by_category}. These categories cover four central capabilities of agents in real-world use cases: Agentic Search (AS, 58 tasks) measures the ability to search for specific pieces of information in complex, heterogeneous scenarios and timelines; Hallucination (HA, 48 tasks) measures the ability to acknowledge what cannot be faithfully answered and to avoid hallucination; Summarization (SU, 14 tasks) measures the ability to gather information across multiple sources; and Circumstance Inference (CI, 20 tasks) measures the ability to comprehend long-horizon communication and discover patterns. Humans manually curated and verified the reference answers and rubrics.

\paragraph{Agentic Search (AS)} Similar to \citet{choubey2025benchmarkingdeepsearchheterogeneous}, this category tests the information-seeking and deep-search capabilities of agents on real-world heterogeneous data. The agent may need to search across multiple Slack channels, GitHub repositories, and Google Docs, as well as external sources on the Internet beyond our environment. Some queries also require agents to correctly track project timeline and GitHub operation timeline. These queries are easily verifiable, typically admitting short answers or clear grading guidelines.

\paragraph{Hallucination (HA)} Inspired by \citet{kirichenko2025abstentionbenchreasoningllmsfail}, we curate queries that test agents in scenarios where they are expected to abstain from answering in order to avoid hallucinating and misleading the user. We focus on three scenarios: (1) \emph{Answer unknown}, e.g., asking for results on a benchmark that was never used in the project; (2) \emph{False premise}, e.g., predicting new results based on a conclusion contrary to what was posted in Slack; (3) \emph{Context restriction}, in which the query specifies a time constraint on Slack history (e.g., before Jan.\ 10th) while the answer can only be found outside this scope. To be judged correct, the agent must not only abstain from answering but also correctly identify the reason for abstention.

\paragraph{Summarization (SU)} We construct queries that ask agents to summarize information toward a given goal, e.g., writing a recommendation letter for a research assistant. Since a research assistant often participates in multiple projects, the agent must search comprehensively across multiple sources to produce a concise and faithful account.

\paragraph{Circumstance Inference (CI)} To further test the ability to comprehend long-horizon conversations, we introduce a novel category, circumstance inference, in which the agent must identify members of the Slack workspace who match a specified circumstance, such as one person acting as a free-rider or two collaborators syncing offline without informing a third. For each query in this category, we construct a dedicated Slack workspace history by modifying the base workspace used in the other three categories according to a predefined reference answer. Concretely, if the reference answer designates Bob as a free-rider, we reassign result-reporting messages to other members, inject additional messages from Bob that use vague or non-committal language, and reduce Bob's participation in substantive task-related threads—thereby embedding the circumstance signal across the conversation history rather than localizing it in any single message. This design requires the agent to synthesize behavioral patterns distributed over long, multi-party conversations in order to arrive at the correct answer.

\section{Experimental Setup}
\label{sec:experiments}

\subsection{Agent Harnesses and Models}

We evaluate 8 agent harnesses across 24 models from frontier labs. For our main experiments, we set the reasoning effort to medium and impose a 20-minute time limit per query. We provide the full list of agents and models in \cref{appendix:agents_models}.

\subsection{Evaluation Setup}
\label{subsec:eval_setup}

All our experiments are conducted using the Harbor framework \citep{Harbor_Framework_Team_Harbor_A_framework_2026}. Specifically, we provide the agents with 2 sets of read-only tools: (1) SlackBench tools that we implement ourselves based on the official Slack Web API methods to search users, list channel history, search messages, etc. (2) GitHub tools to search any publicly available repository. We provide details for each tool in \cref{tab:mcp_tools}. We also enable the terminal read, bash, and web search tools for each agent. We give an illustration of agent trajectory in \cref{fig:trajectory_show}.

\subsection{Answer Verification}

We use LLM judges to evaluate answers. For AS, HA, and CI queries, the answers are short and easy to directly verify in a binary manner. We provide the LLM judge with the query, the agent's answer, the reference answer, and an optional manually curated grading guideline. For Summarization queries, we manually curate checklists for each query, with each checklist item being a key point that must be covered in the summarization. Partial credit is given for each passed checklist item. We use GPT-5.4 as the model for the LLM judge. We provide studies on LLM judge reliability, including human-LLM agreement, in \cref{subsec:llm_judge} and judge prompts in \cref{appendix:prompts}.
\section{Results}
\label{sec:results}

\cref{tab:main_results} reports the accuracy of different agents on each query category, as well as the unweighted average across categories. For each agent and task category, we report the average pass rate across three runs. Antigravity \citep{google2025antigravity} with Gemini 4 Argon \citep{kavukcuoglu2026gemini4argon} achieves the best overall performance, with an average accuracy of 80.25\%, followed by Muse Code with Muse Wonder, Claude Code with Fable 5.1, Claude Code with Opus 5.5, and Claude Code with Muse Wonder. All other agents have overall accuracy under 66\%. These results indicate that complex tasks in realistic and heterogeneous environments can remain challenging even for frontier models and agent harnesses. Performance can vary substantially across categories. For example, OpenHands with GPT-5.5 achieves 60.92\% pass rate on Agentic Search, but shows lower performance than Claude Code with Claude Haiku 4.5 on Hallucination, suggesting that strong agentic search performance does not necessarily translate to reliable recognition of unanswerable or underspecified queries. More generally, we find that all GPT models exhibit weaknesses in the HA category. Further investigation reveals that when asked a question based on a false premise, GPT models tend to first provide a specific answer to the question, only clarifying at the end that this answer rests on a different premise. For example, when asked about the solution status of an OpenRouter issue, GPT 6.1 Sol starts with "Yes—the workaround unblocked the runs," only noting at the end that this answer pertains to the Grok issue, without confirming that no such OpenRouter issue exists. This occurs on 25\% of HA tasks for GPT 6 Astra and 21\% for GPT 6.1 Sol, compared to only 8\% for Opus 5.5 and 6\% for Muse Wonder with Claude Code.

\begin{table}[h]
\centering
\small
\caption{Agent performance on SlackBench. Different categories are defined in \cref{subsec:query_curation}. Overall accuracy is the unweighted average of the 4 categories.}
\label{tab:main_results}
\begin{tabular}{llccccc}
\toprule
Harness & Model & AS & HA & SU & CI & Overall \\
\midrule
Antigravity & Gemini 4 Argon & 70.69 & 90.28 & 76.68 & 83.33 & 80.25 \\
Muse Code & Muse Wonder & 81.03 & 62.50 & 71.42 & 80.00 & 73.74 \\
Claude Code & Claude Fable 5.1 & 67.82 & 71.53 & 71.98 & 83.33 & 73.67 \\
Claude Code & Claude Opus 5.5 & 71.84 & 75.00 & 70.42 & 73.33 & 72.65 \\
Claude Code & Muse Wonder & 68.97 & 66.67 & 67.25 & 80.00 & 70.72 \\
Claude Code & GLM 5.3 & 69.54 & 70.83 & 65.12 & 58.33 & 65.96 \\ 
Grok Build & Grok 4.7 & 64.94 & 56.94 & 70.01 & 70.00 & 65.47 \\
Muse Code & Muse Spark 1.3 & 66.67 & 52.08 & 67.40 & 68.33 & 63.62 \\
Codex CLI & GPT 6.1 Terra & 63.22 & 55.56 & 64.05 & 61.67 & 61.12 \\
Codex CLI & GPT 6 Astra & 62.07 & 54.17 & 65.67 & 60.00 & 60.48 \\
Codex CLI & GPT 6.1 Sol & 59.77 & 50.69 & 62.30 & 60.00 & 58.19 \\
Claude Code & Claude Opus 4.7 & 52.87 & 60.42 & 52.69 & 60.00 & 56.49 \\
OpenHands & GPT-5.5 & 60.92 & 25.69 & 59.30 & 61.67 & 51.89 \\
Terminus-2 & GPT-5.5 & 56.32 & 21.53 & 48.72 & 66.67 & 48.31 \\
Gemini CLI & Gemini 3.1 Pro Preview & 54.02 & 42.36 & 38.74 & 50.00 & 46.28 \\
Claude Code & Claude Sonnet 4.6 & 50.00 & 40.97 & 47.66 & 45.00 & 45.91 \\
OpenHands & Claude Opus 4.6 & 48.28 & 15.28 & 53.24 & 66.67 & 45.87 \\
Terminus-2 & Claude Opus 4.6 & 54.60 & 11.81 & 56.71 & 60.00 & 45.78 \\
Codex CLI & GPT-5.5 & 55.17 & 18.75 & 53.14 & 55.00 & 45.52 \\
Claude Code & Kimi K2.6 & 50.00 & 32.64 & 49.11 & 46.67 & 44.60 \\
Claude Code & Claude Opus 4.6 & 52.87 & 20.83 & 51.39 & 41.67 & 41.69 \\
Terminus-2 & Gemini 3.1 Pro Preview & 50.00 & 34.72 & 27.35 & 50.00 & 40.52 \\
Claude Code & DeepSeek V4 Pro & 43.68 & 37.50 & 47.93 & 30.00 & 39.78 \\
OpenHands & Gemini 3.1 Pro Preview & 44.25 & 31.25 & 36.36 & 43.33 & 38.80 \\
Terminus-2 & Claude Sonnet 4.6 & 51.15 & 15.28 & 44.95 & 43.33 & 38.68 \\
Claude Code & Qwen3.7 Max & 36.78 & 38.19 & 41.23 & 36.67 & 38.22 \\
Claude Code & GLM-5.1 & 50.00 & 22.22 & 50.09 & 30.00 & 38.08 \\
Codex CLI & GPT-5.4 & 51.15 & 21.53 & 40.42 & 36.67 & 37.44 \\
Codex CLI & GPT 6 Luna & 45.98 & 22.92 & 44.05 & 36.67 & 37.40 \\
Gemini CLI & Gemini 3 Flash Preview & 42.53 & 1.39 & 27.26 & 53.33 & 31.13 \\
Claude Code & Claude Haiku 4.5 & 38.51 & 26.39 & 30.15 & 15.00 & 27.51 \\
Codex CLI & GPT-5.4-Mini & 36.21 & 13.89 & 34.73 & 23.33 & 27.04 \\
Terminus-2 & GPT-5.4-Mini & 22.41 & 5.56 & 16.63 & 6.67 & 12.82 \\
\bottomrule
\end{tabular}
\end{table}

\paragraph{Direct CLI access to Slack workspace files improves agent performance and efficiency.}\label{para:cli_vs_mcp} We study the effects of giving agents direct CLI access to all exported Slack workspace files instead of letting them access Slack information through MCP tools, so agents can directly use terminal tools (e.g. grep, jq, cat) to access the workspace. As shown in \cref{tab:no_mcp_results}, across all our evaluated settings, direct CLI access improves pass rates by an average of 7.85\%. Notably, Codex with GPT-5.5 shows 17.25\% increase in accuracy on AS, and slightly outperforms Claude Code with Claude Opus 4.7 overall. We further inspect the agent trajectories for our Claude Code and Codex CLI experiments. We find that CLI access to workspace reduces cost for Claude Code and Opus 5.5 by 54\% and reduces token consumption by 21.00\% on average for the other agents in \cref{tab:cli_vs_mcp}. We include the tool-call composition breakdown in \cref{tab:cli_vs_mcp}. Agents with CLI access perform more direct reading of whole context, while agents with MCP access rely heavily on \texttt{search\_messages}, which returns rich envelopes of metadata that inflate the prompt. 

\begin{table}[h]
\centering
\small
\caption{Performance of agents with MCP tools versus direct terminal access to the Slack workspace. All agents show improved performance with direct terminal access.}
\label{tab:no_mcp_results}
\begin{tabular}{lllccccc}
\toprule
Harness & Model & Access & AS & HA & SU & CI & Overall \\
\midrule
Claude Code & Claude Opus 5.5 & CLI & 66.67 & 72.92 & 70.12 & 85.00 & 73.68 \\
Claude Code & Claude Opus 5.5 & MCP & 71.84 & 75.00 & 70.42 & 73.33 & 72.65 \\
\midrule
Claude Code & Claude Opus 4.7 & CLI & 57.47 & 55.56 & 60.48 & 65.00 & 59.63 \\
Claude Code & Claude Opus 4.7 & MCP & 52.87 & 60.42 & 52.69 & 60.00 & 56.49 \\
\midrule
Claude Code & Claude Sonnet 4.6 & CLI & 58.62 & 41.67 & 55.52 & 63.33 & 54.78 \\
Claude Code & Claude Sonnet 4.6 & MCP & 50.00 & 40.97 & 47.66 & 45.00 & 45.91 \\
\midrule
Codex CLI & GPT-5.5 & CLI & 72.42 & 27.78 & 67.73 & 71.67 & 59.90 \\
Codex CLI & GPT-5.5 & MCP & 55.17 & 18.75 & 53.14 & 55.00 & 45.52 \\
\midrule
Codex CLI & GPT-5.4-Mini & CLI & 53.45 & 18.75 & 48.62 & 51.67 & 43.12 \\
Codex CLI & GPT-5.4-Mini & MCP & 36.21 & 13.89 & 34.73 & 23.33 & 27.04 \\
\midrule
Gemini CLI & Gemini 3.1 Pro Preview & CLI & 62.07 & 37.50 & 51.01 & 66.67 & 54.31 \\
Gemini CLI & Gemini 3.1 Pro Preview & MCP & 54.02 & 42.36 & 38.74 & 50.00 & 46.28 \\
\midrule
Gemini CLI & Gemini 3 Flash Preview & CLI & 42.53 & 3.47 & 33.86 & 58.33 & 34.55 \\
Gemini CLI & Gemini 3 Flash Preview & MCP & 42.53 & 1.39 & 27.26 & 53.33 & 31.13 \\
\bottomrule
\end{tabular}
\end{table}

\begin{table}[h]
  \centering
  \small
  \caption{Tool-call composition for CLI and MCP workspace access, aggregated across runs presented in \cref{tab:no_mcp_results}.}
  \label{tab:cli_vs_mcp}
  \begin{tabular}{lrrrr}
  \toprule
   & \multicolumn{2}{c}{CLI} & \multicolumn{2}{c}{MCP} \\
  \cmidrule(lr){2-3} \cmidrule(lr){4-5}
  Tool-call category & Calls & \% & Calls & \% \\
  \midrule
  Search \& filter (\texttt{grep}/\texttt{jq}/\texttt{awk} vs.\ \texttt{search\_messages}) & 4{,}095 & 30.0\% & 6{,}243 & 50.9\% \\
  Read \& extract (\texttt{cat}/\texttt{head}/\texttt{jq} vs.\ \texttt{get\_channel\_history}) & 7{,}999 & 58.6\% & 3{,}298 & 26.9\% \\
  List \& lookup (\texttt{ls} vs.\ \texttt{list\_channels}, \texttt{get\_user\_info}) & 1{,}164 & 8.5\% & 873 & 7.1\% \\
  Tool / schema loading (built-in vs.\ \texttt{ToolSearch}) & -- & -- & 1{,}848 & 15.1\% \\
  Other (HTTP fetch, etc.) & 393 & 2.9\% & 7 & 0.1\% \\
  \bottomrule
  \end{tabular}
  \end{table}

\paragraph{Higher reasoning effort generally improves agent performance.} As shown in \cref{tab:reasoning_effort}, we study the performance of agents under different reasoning effort settings. We observe a general trend of increase in overall accuracy as we increase the reasoning effort. 

\begin{table}[h]
\centering
\small
\caption{Performance of agents across reasoning effort levels. All agents show increasing performance with higher effort.}
\label{tab:reasoning_effort}
\begin{tabular}{lllccccc}
\toprule
Harness & Model & Effort & AS & HA & SU & CI & Overall \\
\midrule
Claude Code & Claude Opus 4.7 & Low & 50.00 & 50.00 & 45.19 & 50.00 & 48.80 \\
Claude Code & Claude Opus 4.7 & Medium & 52.87 & 60.42 & 52.69 & 60.00 & 56.49 \\
Claude Code & Claude Opus 4.7 & High & 58.62 & 72.92 & 63.94 & 75.00 & 67.62 \\
\midrule
Claude Code & Claude Sonnet 4.6 & Low & 39.66 & 41.67 & 40.86 & 55.00 & 44.30 \\
Claude Code & Claude Sonnet 4.6 & Medium & 50.00 & 40.97 & 47.66 & 45.00 & 45.91 \\
Claude Code & Claude Sonnet 4.6 & High & 55.17 & 33.33 & 44.96 & 60.00 & 48.37 \\
\midrule
Codex CLI & GPT-5.5 & Low & 48.28 & 18.75 & 51.52 & 45.00 & 40.89 \\
Codex CLI & GPT-5.5 & Medium & 55.17 & 18.75 & 53.14 & 55.00 & 45.52 \\
Codex CLI & GPT-5.5 & High & 58.62 & 25.00 & 60.90 & 65.00 & 52.38 \\
\midrule
Codex CLI & GPT-5.4-mini & Low & 25.86 & 10.41 & 24.80 & 0.00 & 15.27 \\
Codex CLI & GPT-5.4-mini & Medium & 36.21 & 13.89 & 34.73 & 23.33 & 27.04 \\
Codex CLI & GPT-5.4-mini & High & 41.38 & 10.42 & 38.09 & 20.00 & 27.47 \\
\midrule
Gemini CLI & Gemini 3.1 Pro Preview & Low & 43.10 & 37.50 & 27.04 & 45.00 & 38.16\\
Gemini CLI & Gemini 3.1 Pro Preview & Medium & 54.02 & 42.36 & 38.74 & 50.00 & 46.28 \\
Gemini CLI & Gemini 3.1 Pro Preview & High & 58.62 & 37.50 & 39.83 & 55.00 & 47.74 \\
\bottomrule
\end{tabular}
\end{table}

\section{Analysis}
\label{sec:analysis}

\subsection{Cost and Statistics}

\begin{figure}[t]
    \centering
    \includegraphics[width=1\columnwidth,height=0.5\columnwidth,keepaspectratio]{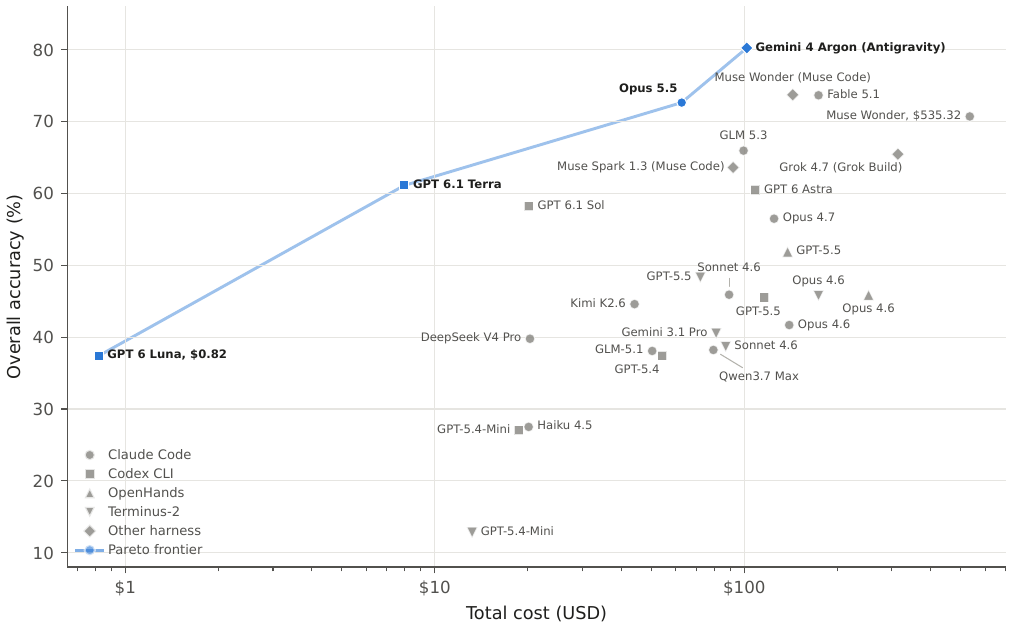}
    \caption{Pareto frontier analysis of SlackBench results, illustrating the trade-off between performance and cost, with cost measured as the average expense of a single run of the entire benchmark.}
    \label{fig:cost_acc}
\end{figure}

\begin{figure}[t]
    \centering
    \includegraphics[width=1\columnwidth,height=0.5\columnwidth,keepaspectratio]{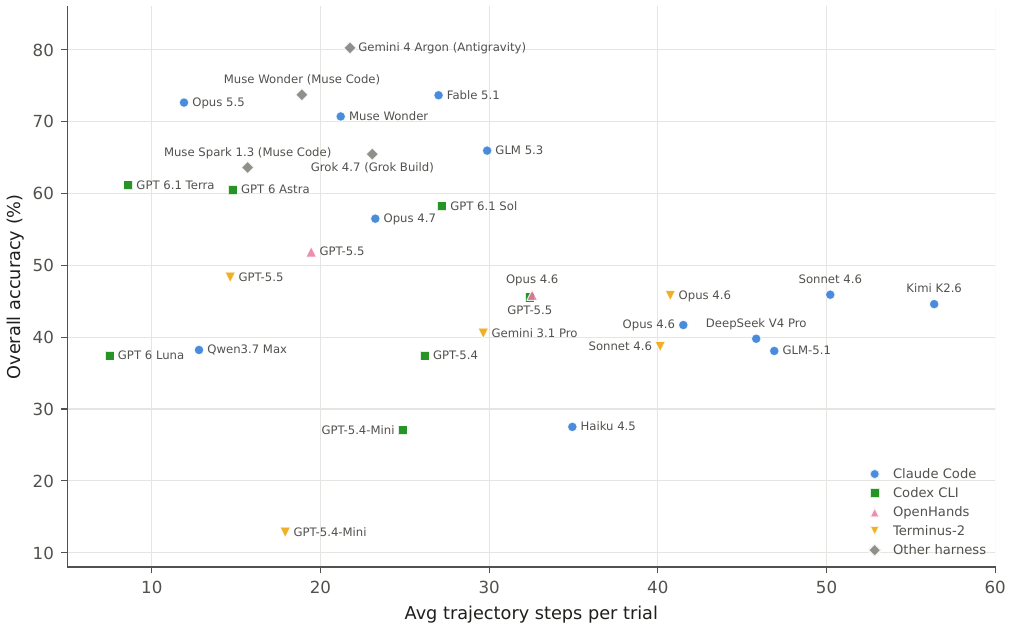}
    \caption{Average trajectory length (number of steps) versus performance for each agent on SlackBench. Trajectory length is measured as the average number of steps per task.}
    \label{fig:steps_acc}
\end{figure}

We show the Pareto frontier analysis and trajectory length analysis in \cref{fig:cost_acc} for agents which we can accurately trace token costs. The cost of running the full SlackBench suite ranges from under \$1 to over \$500, depending on the model's pricing and harness design. Notably, Muse Wonder with Muse Code scores 73.74 with the cost of \$143, while the same model with Claude Code scores 3\% lower and costs more than \$500. However, being closer to the Pareto frontier does not imply more efficient trajectories. We also show each agent's average trajectory length and their pass rate on SlackBench in \cref{fig:steps_acc}. Claude Code with Opus-5.5 takes 11.9 steps on average, while Antigravity with Gemini 4 Argon, though on the Pareto frontier and achieves the highest overall score, takes more than 20 steps on average for each trajectory.

\subsection{Common Failure Modes}

We manually inspect agent trajectories and summarize some common failure modes below, with example trajectories in \cref{appendix:failure_modes}.

\paragraph{Insufficient global verification} Agents often succeed in retrieving plausible candidate answers at first but fail to verify them thoroughly against the broader workspace context. For example, when asked to identify the last meeting between two collaborators, an agent may find a meeting time in one channel and then inspect only nearby messages from that same channel to confirm that it is the last meeting. This local verification misses later evidence in another channel, leading the agent to return an answer that is well supported locally but incorrect globally.

\paragraph{Shallow proxy reasoning} When a query does not have a direct evidence match in Slack or GitHub, agents may fall back on shallow proxies rather than reasoning about the underlying work. For example, when asked to identify the top contributor to a project, an agent may summarize contributions at too high a level and use message volume as a proxy for contribution. This can be misleading: the person sending more messages may simply be running experiments with an existing pipeline, while another contributor performed the harder work of designing or building that pipeline.

\paragraph{Failure to recognize false premises} For hallucination queries containing a false premise, agents can fixate on the incorrect assumption and continue searching under it, often ultimately producing a hallucinated answer.

\paragraph{Confabulation from planning artifacts} When something is discussed but never actually produced, agents often treat planning threads, TODO items, or outline documents as evidence of completion and stitch together a plausible answer from relevant materials. For example, when asked about the contents of a paper's "threats to validity" subsection that was only ever proposed for inclusion, an agent may invent specific disclosures by summarizing nearby analysis files instead of looking for the artifact itself.

\paragraph{Equating absence of keywords with absence of behavior} Some agents treat the absence of literal keyword matches in slack as evidence that the asked-about behavior didn't occur, even when the question requires inference from indirect signals. For example, when asked who on the team has been "reluctant to take on work, often unavailable, or kind of irresponsible," the agent finds no message containing those words and abstains — even though slack contains a clear pattern of one teammate declining tasks, going silent for days, and missing deadlines. The same shallow-proxy logic that produces over-confident wrong answers under loose lexical matching produces unwarranted abstentions under strict lexical matching.

\subsection{Reward Hacking}

During development, we provided agents with a private GitHub token that had access to our SlackBench repository, which contains all queries and reference answers. We found instances where agents navigated directly to this repository and retrieved the reference answers, effectively cheating. Although such instances were rare, this behavior echoes recent findings on evaluation awareness in frontier models \citep{coleman2026evalawareness}. We will not publicly release all the reference answers to our benchmark and will instead make them available to researchers upon request.

\subsection{Comparing mock Slack MCP versus official Slack MCP}
\label{subsec:cmp_mcp_official}

To verify that our custom SlackBench tools reliably approximate the functionality of the official Slack MCP tools (see \cref{subsec:eval_setup}), we evaluate agents on a filtered subset of AS and HA queries using both tool interfaces. We remove queries that can only be answered after our manual filtering of Slack messages (\cref{subsec:env_curation}). As shown in \cref{tab:mcp_comparison}, agent performance with the SlackBench tools is generally close to performance with the official Slack MCP tools. The official Slack MCP tools perform slightly better in some settings, potentially because frontier agents have been better optimized for, or exposed to, these interfaces during development.

\begin{table}[h]
\centering
\small
\caption{Performance comparison between our MCP setup and the official Slack MCP setup.}
\label{tab:mcp_comparison}
\begin{tabular}{lllcc}
\toprule
Harness & Model & Access & AS (filtered) & HA (filtered) \\
\midrule
Claude Code & Claude Opus 4.7 & Ours & 53.33 & 53.19 \\
Claude Code & Claude Opus 4.7 & Official MCP & 57.78 & 59.57 \\
\midrule
Claude Code & Claude Sonnet 4.6 & Ours & 46.67 & 34.04 \\
Claude Code & Claude Sonnet 4.6 & Official MCP & 57.78 & 38.30 \\
\midrule
Codex CLI & GPT-5.5 & Ours & 53.33 & 18.09 \\
Codex CLI & GPT-5.5 & Official MCP & 55.56 & 25.53 \\
\midrule
Codex CLI & GPT-5.4-Mini & Ours & 46.67 & 12.77 \\
Codex CLI & GPT-5.4-Mini & Official MCP & 46.67 & 19.15 \\
\bottomrule
\end{tabular}
\end{table}

\subsection{Human Performance}
\label{subsec:human_performance}

We conducted a human evaluation with 3 independent researchers who were not involved in constructing SlackBench. Humans receive the same environment and tasks as the agents, and are allowed to use AI tools such as Claude Code to complete the tasks. Humans on average achieve 84.5\% accuracy overall, with 91\% on AS, 90\% on HA, 82\% on SU, and 75\% on CI. The 95\% bootstrap confidence interval for overall human accuracy was $[77.0\%, 92.0\%]$. Human performance shows a significant gap between most frontier agents and humans, even on tasks that rely heavily on information retrieval over large collections of artifacts.

\subsection{LLM Judge}
\label{subsec:llm_judge}

To validate the reliability of the LLM judge on SlackBench, we manually review all 264 LLM judgments over all 140 queries for Codex CLI with GPT-5.5 (including rubric judgments for SU tasks). We find that our LLM judge verdicts with GPT-5.4 achieves 95.8\% agreement with human judgment. We further evaluate consistency across different LLM judges by scoring all AS, HA, and CI task results from six agents using GPT-5.4, Claude Opus 4.7, and Gemini 3.1 Pro Preview. As shown in \cref{tab:agreement}, pairwise agreement exceeds 95\% in all cases.

 \begin{table}[h]
\centering
\small
\caption{LLM judge pairwise agreement rate. We use the AS, HA, CI queries from the following 6 agents: Claude Code with Opus 4.7, Claude Code with Sonnet 4.6, Codex CLI with GPT-5.5, Codex CLI with GPT-5.4-Mini, Gemini CLI with Gemini-3.1-Pro-Preview, Gemini CLI with Gemini-3-Flash-Preview.}
\label{tab:agreement}
\begin{tabular}{lccc}
\toprule
& GPT-5.4 & Opus 4.7 & Gemini 3.1 Pro \\
\midrule
GPT-5.4 & 100\% & 95.1\% & 96.3\% \\
Opus 4.7 & -- & 100\% & 96.6\% \\
Gemini 3.1 Pro & -- & -- & 100\% \\
\bottomrule
\end{tabular}
\end{table}
\section{Conclusion}
\label{sec:conclusion}

We introduced SlackBench, a benchmark for evaluating LLM agents on grounded collaborative workspaces that require reasoning across multi-party communication, documents, code, and repository history. SlackBench is constructed through privacy-preserving, human-in-the-loop simulations of academic research workflows, producing realistic Slack conversations tied to GitHub repositories, Google Docs, experiment results, and evolving project decisions. Evaluating 33 frontier models and agent harnesses, we find that even the best system achieves around 80\% overall accuracy, while most agents score less than 66\%, highlighting substantial headroom for current agents in realistic, heterogeneous work environments. Our studies show that interface design strongly affects performance: direct CLI access to exported workspace files improves pass rates and reduces token consumption compared to MCP-based access. Finally, our failure analysis identifies recurring weaknesses in current agents, which can guide future improvements to workspace agents. We release SlackBench as an extensible evaluation framework and invite the community to contribute new environments and queries.

\section{Limitations}
\label{sec:limitations}

For this study, we only run the models and agent harnesses from frontier labs, future work could evaluate more agents to investigate their performance. We also did not test different prompting strategies on agent performance.

\section*{Acknowledgments}
\label{sec:ack}

This project is a collaboration of University of Chicago, UC Berkeley Sky Computing Lab, and Bespoke Labs. This project is partly supported by NSF IFML and Bespoke Labs. Sky Computing Lab is supported by gifts from Accenture, Algorithmic SuperIntelligence Labs, Amazon, AMD, Anyscale, Broadcom, cmpnd, Google, IBM, Intel, Intesa Sanpaolo, Lambda, Lightspeed, Mirendil, NVIDIA, Soma Labs, and VESSL.

\clearpage
\begingroup
\setlength{\bibsep}{1pt plus 0.5pt}
\bibliography{neurips_2026}
\endgroup

\clearpage
\beginsupplement
\crefalias{section}{appendix}
\raggedbottom
\makeatletter
\setlength{\@fptop}{0pt}
\setlength{\@fpsep}{16pt}
\setlength{\@fpbot}{0pt plus 1fil}
\makeatother

\section*{Appendix Contents}
\phantomsection
\label{sec:appendix}

\newcommand{\appendixentry}[2]{%
  \noindent\hspace*{#1}\hyperref[#2]{\ref*{#2}\quad\nameref*{#2}}%
  \nobreak\dotfill\hyperref[#2]{\pageref*{#2}}\par}
\begingroup
\small
\setlength{\parskip}{2pt}
\hypersetup{linkcolor=black}
{\bfseries\appendixentry{0pt}{appendix:chat_gen}}
\smallskip
{\bfseries\appendixentry{0pt}{appendix:failure_modes}}
\smallskip
{\bfseries\appendixentry{0pt}{appendix:prompts}}
\smallskip
{\bfseries\appendixentry{0pt}{appendix:agents_models}}
\endgroup

\section{Details of SlackBench Environment Construction}
\label{appendix:chat_gen}

\subsection{Channel Summaries}

We present content summary for selected channels in \cref{tab:channels}.

\begin{table}[h]
  \centering
  \small
  \caption{Content summary of selected Slack channels.}
  \label{tab:channels}
  \begin{tabular}{ll}
  \toprule
  Channel & Description \\
  \midrule
  bfcl-pv-01 & Prompt variation experiments on the BFCL benchmark \\
  bfcl-thousand-tools & Scaling function-calling evaluation to 16--1024 tools \\
  code-misalignment-capabilities & Running coding and math benchmarks on local models \\
  cognitive-foundations\ldots & Extending a cognitive foundations for reasoning paper \\
  cognitive-paper-code-base & Codebase development for the cognitive paper experiments \\
  hypothesis-literature-agent & Building an agent for hypothesis generation from literature \\
  paper-scraper-project & Scraping ML/NLP conference papers and building a search tool \\
  research-ideagen & Agentic system for automated research ideation from papers \\
  scraper-frontend & Frontend improvements and bug fixes for the paper scraper \\
  tool-bench-langvar & Planning meetings for a tool-use benchmark project \\
  tool-bench-langvar-dev & Development of a multilingual tool-use benchmark \\
  \bottomrule
  \end{tabular}
  \end{table}

\begin{table*}[t]
  \centering
  \scriptsize
  \setlength{\tabcolsep}{5pt}
  \renewcommand{\arraystretch}{1.15}
  \caption{Example queries from each benchmark category.}
  \label{tab:example_queries_by_category}
  \begin{tabular}{p{0.16\textwidth} p{0.79\textwidth}}
  \toprule
  \textbf{Category} & \textbf{Example query} \\
  \midrule

  \textbf{Agentic Search} &
  \begin{tabular}[t]{@{}p{0.79\textwidth}@{}}
  What are the currently remaining TODOs for Ryan in the tool-bench-langvar project?
  \end{tabular}
  \\[2pt]

  \textbf{Hallucination} &
  \begin{tabular}[t]{@{}p{0.79\textwidth}@{}}
  I remember that for our prompt variation project for BFCL, changing return format to typescript + tool call tags would decrease the performance for gpt-4o, could you find how much the decrease is?
  \end{tabular}
  \\[2pt]

  \textbf{Summarization} &
  \begin{tabular}[t]{@{}p{0.79\textwidth}@{}}
  I'm trying to write a recommendation letter for Marcus for his PhD application. Could you summarize his research contributions?
  \end{tabular}
  \\[2pt]

  \textbf{Circumstance inference} &
  \begin{tabular}[t]{@{}p{0.79\textwidth}@{}}
  one of our team keeps taking credit for someone else's work. who's the credit-stealer and who's getting taken from?
  \end{tabular}
  \\

  \bottomrule
  \end{tabular}
  \end{table*}

\subsection{Tool Description}

We present SlackBench and GitHub tools provided to agents in \cref{tab:mcp_tools}. For Claude Code, Codex CLI, Gemini CLI, Antigravity CLI, Grok Build, and Muse Code, we provide these tools via MCP servers. For Terminus-2 and OpenHands, we provide equivalent functionality through a command-line tool that wraps the same underlying API endpoints, as these agents primarily interact through terminal commands rather than MCP. 

\begin{table}[h]
  \centering
  \small
  \caption{MCP tools available to agents. Slack tools provide read-only access to workspace messages, channels, and users. GitHub tools provide read-only access to repositories,
  code, issues, and pull requests.}
  \label{tab:mcp_tools}
  \begin{tabular}{ll}
  \toprule
  Tool & Description \\
  \midrule
  \multicolumn{2}{l}{\textit{Slack Tools (7)}} \\
  \texttt{list\_channels} & Lists all channels in the Slack workspace \\
  \texttt{get\_channel\_history} & Fetches message history for a channel \\
  \texttt{get\_channel\_members} & Lists members of a channel \\
  \texttt{get\_thread\_replies} & Fetches replies in a message thread \\
  \texttt{get\_user\_info} & Gets information about a user by ID \\
  \texttt{list\_users} & Lists all users in the workspace \\
  \texttt{search\_messages} & Searches messages by query \\
  \midrule
  \multicolumn{2}{l}{\textit{GitHub Tools (14, read-only)}} \\
  \texttt{search\_repositories} & Searches for repositories \\
  \texttt{search\_code} & Searches for code across repositories \\
  \texttt{search\_issues} & Searches for issues and pull requests \\
  \texttt{search\_users} & Searches for users \\
  \texttt{get\_file\_contents} & Gets contents of a file in a repository \\
  \texttt{get\_issue} & Gets details of a specific issue \\
  \texttt{get\_pull\_request} & Gets details of a pull request \\
  \texttt{get\_pull\_request\_files} & Gets files changed in a pull request \\
  \texttt{get\_pull\_request\_status} & Gets status/checks of a pull request \\
  \texttt{get\_pull\_request\_comments} & Gets comments on a pull request \\
  \texttt{get\_pull\_request\_reviews} & Gets reviews on a pull request \\
  \texttt{list\_commits} & Lists commits in a repository \\
  \texttt{list\_issues} & Lists issues in a repository \\
  \texttt{list\_pull\_requests} & Lists pull requests in a repository \\
  \bottomrule
  \end{tabular}
  \end{table}
\section{Example Failure Trajectories}
\label{appendix:failure_modes}

\subsection{Example 1}
\label{appendix:failure_modes:insufficient_search}

\textbf{Query.}
  \begin{verbatim}
  What is the title of the paper that Marcus referred to in the tool-bench-langvar
  project channel when Andy and Brandon talked to him?
  \end{verbatim}

  \textbf{Final answer.}
  \begin{verbatim}
  $\tau^2$-Bench: Evaluating Conversational Agents in a Dual-Control Environment
  \end{verbatim}

  \textbf{Reference answer.}
  \begin{verbatim}
  Persona Features Control Emergent Misalignment
  (need to have the exact title, capitalization does not matter)
  \end{verbatim}

  \textbf{Judge result.} \texttt{FAIL (llm\_judge, reward=0.000)}.

  {\footnotesize
  \begin{enumerate}
  \item[\textbf{Step 1.}] The user asks for the title of the paper Marcus referred to when Andy and Brandon talked to him in the tool-bench-langvar project channel.
  \item[\textbf{Step 2.}] The agent lists channels and finds the relevant langvar channels, including \texttt{tool-bench-langvar} and \texttt{tool-bench-langvar-dev}.
  \item[\textbf{Step 3.}] It loads the history of \texttt{tool-bench-langvar}; the response contains ordinary langvar project updates, but no obvious paper title.
  \item[\textbf{Step 4.}] It lists all users to map names to user IDs.
  \item[\textbf{Step 5.}] It lists users again, repeating the identity-resolution step.
  \item[\textbf{Step 6.}] It searches for ``Marcus'' globally; this returns many Marcus-related messages across channels, but no exact paper title.
  \item[\textbf{Step 7.}] It searches for ``paper channel:tool-bench-langvar''; the search returns no hits.
  \item[\textbf{Step 8.}] It searches for ``Marcus paper''; the search returns no hits.
  \item[\textbf{Step 9.}] It searches for ``Marcus Andy Brandon''; the search returns no hits.
  \item[\textbf{Step 10.}] It loads the history of \texttt{tool-bench-langvar-dev}; this returns a large dev-channel dump containing many experiment/result messages.
  \item[\textbf{Step 11.}] It fetches user info for Marcus and confirms that Marcus corresponds to user ID \texttt{U0A7ZS0PR50}.
  \item[\textbf{Step 12.}] It reloads \texttt{tool-bench-langvar-dev} history, again getting a large dump of recent dev-channel messages.
  \item[\textbf{Step 13.}] It searches for ``from:@Marcus Sullivan paper''; the search returns no hits.
  \item[\textbf{Step 14.}] It searches for ``from:U0A7ZS0PR50''; this returns Marcus's messages, but still not an explicit paper title.
  \item[\textbf{Step 15.}] It reloads \texttt{tool-bench-langvar-dev} history once more.
  \item[\textbf{Step 16.}] It reloads \texttt{tool-bench-langvar-dev} history yet again.
  \item[\textbf{Step 17.}] It searches for ``paper'' globally; this returns a large noisy set of paper mentions across the workspace.
  \item[\textbf{Step 18.}] It loads a short recent slice of \texttt{tool-bench-langvar-dev} history (limit 20); still no exact title is extracted.
  \item[\textbf{Step 19.}] It loads a larger recent slice of \texttt{tool-bench-langvar-dev} history (limit 100); still no exact title is extracted.
  \item[\textbf{Step 20.}] It searches for ``EM paper''; this is the key retrieval step, and it finds Marcus's message: ``reading the EM paper rn, will get results later.''
  \item[\textbf{Step 21.}] It searches for ``Marcus title''; the search returns no hits.
  \item[\textbf{Step 22.}] It searches for ``Marcus from:U09PNDT1W9E'' to connect Marcus with Andy; the search returns no hits.
  \item[\textbf{Step 23.}] It searches for ``Marcus from:elliot''; the search returns no hits.
  \item[\textbf{Step 24.}] It searches for ``Brandon Marcus''; this returns direct Marcus--Brandon interaction messages, but not the paper title.
  \item[\textbf{Step 25.}] It searches for ``Marcus channel:tool-bench-langvar''; the search returns no hits.
  \item[\textbf{Step 26.}] It searches for ``EM''; this returns broad EM-related matches, including the same ``EM paper'' clue.
  \item[\textbf{Step 27.}] It fetches user info for Brandon Kong, checking whether this is the relevant Brandon.
  \item[\textbf{Step 28.}] It searches for messages from Brandon Kong or Andy in \texttt{tool-bench-langvar}; the search returns no hits.
  \item[\textbf{Step 29.}] It searches for ``Brandon channel:tool-bench-langvar''; the search returns no hits.
  \item[\textbf{Step 30.}] It reloads \texttt{tool-bench-langvar-dev} history again.
  \item[\textbf{Step 31.}] It fetches user info for Ryan, presumably to disambiguate participants mentioned in nearby messages.
  \item[\textbf{Step 32.}] It searches for ``Brandon paper''; the search returns no hits.
  \item[\textbf{Step 33.}] It searches for ``Andy paper''; the search returns no hits.
  \item[\textbf{Step 34.}] It searches for the two user IDs of Andy and Brandon Kong together; the search returns no hits.
  \item[\textbf{Step 35.}] It lists channels again, repeating earlier channel discovery.
  \item[\textbf{Step 36.}] It opens the root of the GitHub repository \texttt{tool-bench-langvar-writing}; the response shows repository contents.
  \item[\textbf{Step 37.}] It tries to read a local \texttt{README.md}; this does not add useful information for the paper-identification task.
  \item[\textbf{Step 38.}] It fetches \texttt{README.md} from the GitHub writing repo; this still does not identify the paper title.
  \item[\textbf{Step 39.}] It opens the \texttt{results} directory of the writing repo; this reveals result files, but these are not directly relevant to Marcus's paper reference.
  \item[\textbf{Step 40.}] It loads the history of \texttt{misalign-capabilities}; this is the correct cross-channel direction, but the agent does not yet extract the
  exact paper title.
  \item[\textbf{Step 41.}] It reloads \texttt{tool-bench-langvar-dev} history, returning to the local langvar context.
  \item[\textbf{Step 42.}] It searches for ``Marcus channel:all-icetea-ai-research''; the search returns no hits.
  \item[\textbf{Step 43.}] It searches for ``referred to'' globally; the results are noisy and not decisive.
  \item[\textbf{Step 44.}] It searches for ``Easy-to-Misalign''; the search returns no hits.
  \item[\textbf{Step 45.}] It loads another recent slice of \texttt{tool-bench-langvar-dev} history (limit 50); again, no exact title is extracted.
  \item[\textbf{Step 46.}] It reloads \texttt{tool-bench-langvar} history, revisiting the main langvar channel.
  \item[\textbf{Step 47.}] It reloads \texttt{tool-bench-langvar-dev} history with limit 100.
  \item[\textbf{Step 48.}] It opens \texttt{results/complexfuncbench\_results.md} from the writing repo; this is another irrelevant detour into result files.
  \item[\textbf{Step 49.}] It searches for messages from Brandon Kong together with Marcus; the search returns no hits.
  \item[\textbf{Step 50.}] It searches for ``Marcus Brandon'' again; this returns interaction context, but still not the paper title.
  \item[\textbf{Step 51.}] It reloads \texttt{tool-bench-langvar-dev} history yet again.
  \item[\textbf{Step 52.}] It searches for ``paper channel:tool-bench-langvar-dev''; the search returns no hits.
  \item[\textbf{Step 53.}] It searches for ``EM paper'' again; this re-finds Marcus's ``reading the EM paper rn'' message.
  \item[\textbf{Step 54.}] It searches for ``EM paper title''; the search returns no hits.
  \item[\textbf{Step 55.}] It searches for ``Marcus paper'' again; the search returns no hits.
  \item[\textbf{Step 56.}] It searches for ``referred to'' again; this repeats the earlier noisy search.
  \item[\textbf{Step 57.}] It searches for ``Marcus Brandon'' again; this repeats the earlier interaction search.
  \item[\textbf{Step 58.}] It fetches user info for Henry Moss, another participant in the langvar channels, but this does not help identify the paper.
  \item[\textbf{Step 59.}] It searches for ``paper channel:bfcl-more-tools''; the search returns no hits and is unrelated to the target context.
  \item[\textbf{Step 60.}] It searches for ``Marcus Sullivan paper''; the search returns no hits.
  \item[\textbf{Step 61.}] It searches for ``tool-bench-langvar'' globally; this returns project/repo messages, but still not the paper title.
  \item[\textbf{Step 62.}] It searches for ``EchoMimic'' as one possible expansion of ``EM''; the search returns no hits.
  \item[\textbf{Step 63.}] It reloads \texttt{tool-bench-langvar-dev} history with limit 100.
  \item[\textbf{Step 64.}] It searches for ``Effective Multilingual paper'' as another speculative expansion of ``EM''; the search returns no hits.
  \item[\textbf{Step 65.}] It searches for ``EM channel:tool-bench-langvar-dev''; the search returns no hits.
  \item[\textbf{Step 66.}] It searches for ``EM'' again; this returns broad EM-related matches, including the same Marcus message.
  \item[\textbf{Step 67.}] It searches for ``confound''; the results are unrelated to the paper title.
  \item[\textbf{Step 68.}] It searches for ``Marcus referred to paper''; the search returns no hits.
  \item[\textbf{Step 69.}] It searches for ``Marcus paper Andy''; the search returns no hits.
  \item[\textbf{Step 70.}] It searches for ``Marcus paper Brandon''; the search returns no hits.
  \item[\textbf{Step 71.}] It searches for ``EM paper'' a third time; this again recovers the same clue but does not disambiguate it.
  \item[\textbf{Step 72.}] It reloads \texttt{tool-bench-langvar-dev} history yet again.
  \item[\textbf{Step 73.}] It loads \texttt{misalign-capabilities} history with limit 100; this is the key channel that contains the correct full title, but the agent
  still does not extract it.
  \item[\textbf{Step 74.}] It reloads \texttt{tool-bench-langvar} history again.
  \item[\textbf{Step 75.}] It reloads \texttt{tool-bench-langvar-dev} history again.
  \item[\textbf{Step 76.}] It reloads \texttt{tool-bench-langvar-dev} history with limit 100 again.
  \item[\textbf{Step 77.}] It reloads \texttt{tool-bench-langvar-dev} history with limit 100 yet again.
  \item[\textbf{Step 78.}] It searches for ``paper multilingual Marcus''; the search returns no hits.
  \item[\textbf{Step 79.}] It searches for ``paper Marcus tau2''; the search returns no hits.
  \item[\textbf{Step 80.}] It reloads \texttt{tool-bench-langvar-dev} history with limit 100 once more.
  \item[\textbf{Step 81.}] It lists channels again, repeating earlier channel discovery.
  \item[\textbf{Step 82.}] It searches for ``Brandon asked''; this returns a message about Marcus posting model results, not about the paper title.
  \item[\textbf{Step 83.}] It searches for ``Marcus confound''; the search returns no hits.
  \item[\textbf{Step 84.}] It searches for ``Marcus user simulator''; the search returns no hits.
  \item[\textbf{Step 85.}] It searches for ``Marcus language''; the search returns no hits.
  \item[\textbf{Step 86.}] It searches for messages from Brandon Kong containing ``paper''; the search returns no hits.
  \item[\textbf{Step 87.}] It searches for messages from Andy containing ``paper''; the search returns no hits.
  \item[\textbf{Step 88.}] It searches for ``Marcus Wang''; the search returns no hits.
  \item[\textbf{Step 89.}] It searches for ``Marcus self-consistency''; the search returns no hits.
  \item[\textbf{Step 90.}] It searches for ``Marcus reasoning''; the search returns no hits.
  \item[\textbf{Step 91.}] It searches for ``Marcus foundation''; the search returns no hits.
  \item[\textbf{Step 92.}] It searches for ``The Effect of Language on Function Calling''; the search returns no hits.
  \item[\textbf{Step 93.}] It gives the final answer that the paper was \texttt{$\tau^2$-Bench: Evaluating Conversational Agents in a Dual-Control Environment}, explicitly saying
  the broader discussion centered on that paper.
  \end{enumerate}
  }

  \textbf{Failure analysis.}
  This trajectory is an example of \emph{insufficient search after partial retrieval}. The agent successfully retrieves the crucial intermediate clue---Marcus's message
  ``reading the EM paper rn'' (Steps 20, 53, and 71)---but never properly resolves what ``EM'' refers to. Instead, it repeatedly loops through local langvar searches, speculative
  acronym expansions (e.g., ``EchoMimic'' and ``Effective Multilingual''), and irrelevant detours into writing-repo files. Although it does eventually open the correct cross-
  channel context in \texttt{misalign-capabilities} (Steps 40 and 73), it still fails to extract the exact title \texttt{Persona Features Control Emergent Misalignment}.
  Because \texttt{tool-bench-langvar-dev} already contains a highly salient explicit paper mention for \texttt{$\tau^2$-Bench}, the agent falls back to that nearby titled paper
  rather than disambiguating the shorthand reference across channels.

\subsection{Example 2}
\label{appendix:failure_modes:lack_verify}

\textbf{Query.}
  \begin{verbatim}
  When testing the performance of BFCL with both change in the number of available
  functions and prompt variation, did the researcher use the newer version of BFCL
  or the older version?
  \end{verbatim}

  \textbf{Final answer.}
  \begin{verbatim}
  The researcher used the older BFCL setup, not the newer official version. In
  Slack they noted that the current BFCL version did not have the --prompt-variation
  flag, then pointed to the older bfcl-prompt-v-01 prompt-variation project as the source
  of that functionality and ran the scaling/prompt-variation tests from that older
  setup (later falling back to manual prompt scripts for part of the experiment).
  \end{verbatim}

  \textbf{Reference answer.}
  \begin{verbatim}
  The newer version of BFCL
  \end{verbatim}

  \textbf{Judge result.} \texttt{FAIL (llm\_judge, reward=0.000)}.

  {\footnotesize
  \begin{enumerate}
  \item[\textbf{Step 1.}] The system prompt injects tools, permissions, and environment policy.
  \item[\textbf{Step 2.}] The environment context is injected (\texttt{/app}, \texttt{bash}, date \texttt{2026-03-30}, timezone \texttt{UTC}).
  \item[\textbf{Step 3.}] The user provides the BFCL-version query and requires the answer in \texttt{FINAL\_ANSWER:} format.
  \item[\textbf{Step 4.}] The agent says it will inspect Slack for mentions of BFCL, prompt variation, and changes in the number of available functions to determine which BFCL
  version was used.
  \item[\textbf{Step 5.}] It searches for \texttt{BFCL prompt variation number of available functions newer older version}; the search returns no exact match.
  \item[\textbf{Step 6.}] It searches for \texttt{BFCL available functions prompt variation}; this also fails to directly answer the question.
  \item[\textbf{Step 7.}] It searches for \texttt{BFCL newer version older version}; again, no direct answer is found.
  \item[\textbf{Step 8.}] The agent notes that the first pass was too narrow and decides to broaden to more general BFCL experiment-setup discussion.
  \item[\textbf{Step 9.}] It searches for \texttt{BFCL}; this surfaces broad BFCL-related discussion and points it toward the relevant project area.
  \item[\textbf{Step 10.}] It searches for \texttt{"function calling" BFCL}; this yields discussion around BFCL tooling and setup.
  \item[\textbf{Step 11.}] It searches for \texttt{"available functions" researcher prompt variation}; this does not directly resolve the version question.
  \item[\textbf{Step 12.}] The agent says it has found the relevant thread area and now wants to determine whether the team used the original BFCL fork or a newer release.
  \item[\textbf{Step 13.}] It searches for \texttt{"prompt variation" BFCL fork modified version original BFCL}; the result is not decisive.
  \item[\textbf{Step 14.}] It searches for \texttt{"for now" "original bfcl" prompt variation}; again, no decisive answer.
  \item[\textbf{Step 15.}] It searches for \texttt{"--prompt-variation" BFCL version}; this retrieves the key early message that the current BFCL version does not have the
  \texttt{--prompt-variation} flag.
  \item[\textbf{Step 16.}] It searches for \texttt{"previous project" BFCL prompt variation older}; this is still probing whether the feature only existed in the older setup.
  \item[\textbf{Step 17.}] It searches for \texttt{"not in official BFCL" previous project BFCL}; this retrieves the key early message saying the feature is not in official BFCL
  and pointing to the previous project \texttt{bfcl-prompt-v-01}.
  \item[\textbf{Step 18.}] It searches for \texttt{"our previous project" "bfcl-prompt-v-01"}; this reinforces the same older-project evidence.
  \item[\textbf{Step 19.}] The agent says it now has enough to answer tentatively, but wants to inspect neighboring messages in the same channel in case the team later switched
  back to the newer codebase.
  \item[\textbf{Step 20.}] It loads recent history from the \texttt{bfcl-more-tools} channel (\texttt{limit=40}). This is the critical verification step: the retrieved context
  already contains later evidence that prompt-variation support was integrated into the current project and that the actual combined experiments were run there.
  \item[\textbf{Step 21.}] Despite having fetched the later context, the agent does not revise its hypothesis and gives the final answer that the older BFCL setup was used.
  \end{enumerate}
  }

  \textbf{Failure analysis.}
  This is a clean example of \emph{lack of context-based answer verification}. The agent retrieves a plausible early answer: the current BFCL version initially lacked
  \texttt{--prompt-variation}, and the older \texttt{bfcl-prompt-v-01} project already had that functionality. However, instead of verifying whether this remained true for the final
  experiment setup, it anchors on that early evidence and stops updating its belief. Crucially, the later channel context it fetches in Step~20 already contains the disambiguating
  evidence: the team explicitly says they are integrating the prompt-variation code into the current \texttt{bfcl-more-tools} project (\texttt{2026-01-15.json:573}), then says
  they \emph{added} the \texttt{--prompt-variation} flag to the CLI and are starting the actual experiments (\texttt{2026-01-15.json:612}), and later reports running the combined
  prompt-variation experiments at both pool~0 and pool~512 (\texttt{2026-01-15.json:1401}, \texttt{2026-01-15.json:1490}, \texttt{2026-01-15.json:2036}). The failure is therefore
  not a simple retrieval miss: the agent finds both the early misleading evidence and the later corrective evidence, but fails to use the later context to verify and overturn its
  initial answer.

\subsection{Example 3}
\label{appendix:failure_modes:false_premise}

\textbf{Query.}
  \begin{verbatim}
  In the research-ideagen channel, Jordan recommended simplifying the experiment
  from 3 conditions to 2 conditions. What was the third condition that was removed?
  \end{verbatim}

  \textbf{Final answer.}
  \begin{verbatim}
  The third condition that was removed was the "collaborative" condition. Jordan
  recommended simplifying the experiment to just two conditions — control (no AI)
  vs. AI-assisted — by skipping the collaborative condition, as it would be easier
  to recruit participants and produce cleaner results.
  \end{verbatim}

  \textbf{Reference answer.}
  \begin{verbatim}
  The answer must explicitly point out that it is Nancy, NOT Jordan, who made
  the recommendation. Otherwise, it is incorrect. The answer can provide a definite
  answer, but it needs to be: the third condition was 'AI-collaborative.'.
  Otherwise, it is incorrect. (i.e. abstain from answering OR providing the correct
  answer are both correct, but it needs to point out that it is Nancy, NOT Jordan,
  who made the recommendation)
  \end{verbatim}

  \textbf{Judge result.} \texttt{FAIL (llm\_judge, reward=0.000)}.

  {\footnotesize
  \begin{enumerate}
  \item[\textbf{Step 1.}] The system prompt injects the tool environment and run policy.
  \item[\textbf{Step 2.}] The environment context is injected.
  \item[\textbf{Step 3.}] The user provides the abstention query and required answer format.
  \item[\textbf{Step 4.}] The agent begins by searching the \texttt{research-ideagen} channel for discussion of simplifying the experiment from three conditions to two.
  \item[\textbf{Step 5.}] It issues initial targeted searches for the simplification discussion, focusing on Jordan, experiment conditions, and the relevant channel.
  \item[\textbf{Step 6.}] These initial searches do not directly resolve the question, so the agent broadens to adjacent experiment-design discussion in the same channel.
  \item[\textbf{Step 7.}] It pulls channel history and nearby messages around the experiment-design thread.
  \item[\textbf{Step 8.}] It extracts the relevant cluster of messages discussing control, AI-assisted, and collaborative conditions.
  \item[\textbf{Step 9.}] It finds the key message from Nancy saying to skip the collaborative condition and simplify to control versus AI-assisted only.
  \item[\textbf{Step 10.}] It also finds Jordan's follow-up message agreeing with the simplification and restating the two-condition setup.
  \item[\textbf{Step 11.}] At this point, the agent explicitly notices the contradiction: Nancy is the one who actually says to skip the collaborative condition, while Jordan
  only confirms the plan afterward.
  \item[\textbf{Step 12.}] Instead of treating this as a false-premise correction, the agent reasons that Jordan's confirmation is close enough to count as the recommendation.
  \item[\textbf{Step 13.}] It continues reasoning under the user-provided premise that Jordan was the recommender.
  \item[\textbf{Step 14.}] It identifies the removed condition as the collaborative condition.
  \item[\textbf{Step 15.}] It states, in effect, that ``either way'' the removed condition was collaborative, thereby collapsing the distinction between who made the
  recommendation and what the removed condition was.
  \item[\textbf{Step 16.}] It gives the final answer that Jordan recommended dropping the collaborative condition.
  \end{enumerate}
  }

  \textbf{Failure analysis.}
  This is a direct example of \emph{failure to recognize a false premise}. The agent successfully retrieves the contradictory evidence and even articulates it internally: Nancy,
  not Jordan, is the speaker who explicitly says to skip the collaborative condition. However, instead of correcting the user’s premise, it continues answering as though the
  premise were effectively true because Jordan later agreed with the simplification. The benchmark is designed so that recognizing the wrong attribution is essential: either the
  agent must explicitly correct the premise (``Nancy, not Jordan'') or abstain. Instead, it preserves the false premise and only answers the substantive sub-question about which
  condition was removed.

\subsection{Example 4}
\label{appendix:failure_modes:polished_narrative}

\textbf{Query.}
  \begin{verbatim}
  I'm trying to write a recommendation letter for Marcus for his PhD application.
  Could you summarize his research contributions?
  \end{verbatim}

  \textbf{Final answer.}
  \begin{verbatim}
  Based on the workspace history, Marcus Sullivan’s strongest research contributions
  are in empirical evaluation of LLM tool use. He appears to have led or substantially
  driven a large-scale BFCL “function calling at scale” study, producing comprehensive
  results across 5 models, 3 task categories, and tool-pool sizes up to 512 (105 total
  evaluations), along with polished comparison plots and written analyses. That work
  surfaced several actionable findings: Grok-3-beta was the strongest overall model,
  sequential multi-call tasks scaled much better than mixed parallel/sequential tasks,
  mini models degraded sharply beyond about 128-256 tools, and context window size
  alone did not explain performance. He also contributed to multilingual robustness
  research, running and analyzing language-variation experiments on ComplexFuncBench
  and tau2-bench. In those studies he helped establish that language effects are
  benchmark-dependent, that Spanish consistently underperformed despite being a
  high-resource language, that stronger models reduce cross-lingual performance gaps,
  and that some apparent language effects may actually come from user-simulator
  confounds rather than model capability alone. Finally, he seems to have contributed
  beyond raw experimentation by debugging evaluation infrastructure, running difficult
  model sweeps, organizing results for collaborators, and turning findings into
  repo-ready analyses and draft writing material.
  \end{verbatim}

  \textbf{Reference answer.}
  \begin{verbatim}
  Checklist-based grading. The summary must mention Marcus's contributions in exactly
  these projects: tool-bench-langvar, bfcl-more-tools, and
  misalign-capabilities, with tightly scoped contributions in each.
  For example, for tool-bench-langvar it should only mention that Marcus tested
  tau2-bench-retail on language variations; for bfcl-more-tools it should only
  mention that Marcus mainly ran the benchmark built by others and analyzed results;
  for misalign-capabilities it should only mention that Marcus helped Andy
  find benchmarks, run SWE-bench, do literature review, and create HumanEval-based
  correct/incorrect code mixtures.
  \end{verbatim}

  \textbf{Judge result.} Checklist score below full credit; see
  \texttt{verifier/checklist\_results.txt} in the run directory.

  {\footnotesize
  \begin{enumerate}
  \item[\textbf{Step 1.}] The system prompt injects tools, permissions, and environment policy.
  \item[\textbf{Step 2.}] The environment context is injected.
  \item[\textbf{Step 3.}] The user asks for a summary of Marcus's research contributions for a recommendation letter.
  \item[\textbf{Step 4.}] The agent starts by identifying channels and repositories associated with Marcus's research projects.
  \item[\textbf{Step 5.}] It searches Slack for Marcus's name and retrieves broad references across multiple projects.
  \item[\textbf{Step 6.}] It inspects the \texttt{tool-bench-langvar} and \texttt{tool-bench-langvar-dev} channels, pulling evidence about Marcus's language-variation work.
  \item[\textbf{Step 7.}] It inspects BFCL-related channels, especially \texttt{bfcl-more-tools}, collecting messages about Marcus running evaluations, generating plots, and
  discussing results.
  \item[\textbf{Step 8.}] It inspects the \texttt{misalign-capabilities} channel and gathers messages showing Marcus helping with benchmarks, SWE-bench, literature
  review, and dataset construction.
  \item[\textbf{Step 9.}] It expands the search into GitHub repositories and writing repos associated with these projects.
  \item[\textbf{Step 10.}] It reads project artifacts and commit histories to infer the broader story of Marcus's involvement.
  \item[\textbf{Step 11.}] It synthesizes the retrieved evidence into a high-level picture of Marcus as an empirical LLM evaluation researcher.
  \item[\textbf{Step 12.}] Instead of preserving the narrow scope of each contribution, the agent begins to aggregate them into broad themes such as empirical evaluation,
  multilingual robustness, and infrastructure/debugging.
  \item[\textbf{Step 13.}] It upgrades bounded contributions into stronger claims, e.g., Marcus ``led or substantially driven'' the BFCL scaling study.
  \item[\textbf{Step 14.}] It broadens the tool-bench-langvar contribution from a specific tau2-retail experiment to a general multilingual robustness research narrative spanning
  ComplexFuncBench and tau2-bench.
  \item[\textbf{Step 15.}] It broadens the BFCL contribution from running an existing benchmark pipeline and analyzing results to an overarching leadership-style description of
  the project and its findings.
  \item[\textbf{Step 16.}] It broadens the misalign-capabilities contribution from a scoped support role to a more expansive description of research and infrastructure
  work.
  \item[\textbf{Step 17.}] It composes the final answer as polished recommendation-style prose emphasizing importance, impact, and thematic coherence.
  \item[\textbf{Step 18.}] It returns the final summary without checking whether the wording stays within the checklist's allowed contribution boundaries.
  \end{enumerate}
  }

  \textbf{Failure analysis.}
  This is a clear example of \emph{polished narrative instead of detail-grounded summary}. The agent retrieves the right general evidence and identifies the correct projects, but
  it does not behave like an evidence-constrained summarizer. Instead, it writes as if it were already drafting the recommendation letter itself, producing fluent advocacy-style
  prose that merges distinct contributions into broad research themes. In doing so, it inflates Marcus's role (e.g., from running and analyzing experiments to ``led or
  substantially driven''), introduces contributions outside the allowed scope for particular projects, and omits the fine-grained project-by-project boundaries required by the
  checklist. The failure is therefore not lack of retrieval, but loss of fidelity during summarization: the answer is polished and plausible, yet insufficiently grounded in the
  exact underlying evidence.

\subsection{Example 5}
\label{appendix:failure_modes:premature_closure}

\textbf{Query.}
  \begin{verbatim}
  I'm reviewing our peer feedback culture. Someone flagged a concern about one team
  member who seems to go beyond normal code review — consistently dismissing or
  undermining another specific person's results in a way that feels targeted rather
  than constructive. (Project lead questioning RA does not count.) Can you find who
  is doing this and who they're targeting? I need both names.
  \end{verbatim}

  \textbf{Final answer.}
  \begin{verbatim}
  Chris (U0A6ZART83A) is the one doing the dismissing, and Alex (U0A62TFKX38)
  is the one being targeted.
  \end{verbatim}

  \textbf{Reference answer.}
  \begin{verbatim}
  Ryan doubting Marcus
  \end{verbatim}

  \textbf{Judge result.} \texttt{FAIL (llm\_judge, reward=0.000)}.

  {\footnotesize
  \begin{enumerate}
  \item[\textbf{Step 1.}] The user asks who is repeatedly dismissing another peer's results, with the explicit note that project-lead-to-RA criticism does not count.
  \item[\textbf{Step 2.}] The agent states that it will search Slack for cases where one person consistently undermines another person's work.
  \item[\textbf{Step 3.}] It uses tool search to select \texttt{list\_channels} and \texttt{search\_messages}.
  \item[\textbf{Step 4.}] It lists workspace channels.
  \item[\textbf{5--9.}] It issues broad searches such as \texttt{dismissing undermining results}, \texttt{code review negative feedback targeted}, \texttt{wrong flawed
  incorrect results}, \texttt{don't trust work sloppy}, and \texttt{disagree criticism doubt questionable}. None of these directly identify a pair.
  \item[\textbf{10--15.}] After the keyword searches fail, the agent explicitly switches to direct channel-history inspection.
  \item[\textbf{Step 16.}] It uses tool search to select \texttt{get\_channel\_history} and \texttt{list\_users}.
  \item[\textbf{Step 17.}] It lists all users so it can map names to IDs.
  \item[\textbf{18--20.}] It loads history for three major channels: \texttt{cognitive-foundations-for-reasoning-paper-extension}, \texttt{bfcl-more-tools}, and
  \texttt{misalign-capabilities}.
  \item[\textbf{21--23.}] It launches helper agents to inspect those large channel histories.
  \item[\textbf{24--28.}] It checks additional channels, including \texttt{research-ideagen}, \texttt{paper-operator-project}, \texttt{operator-frontend}, and another smaller
  channel.
  \item[\textbf{29--32.}] It states that some of those channels did not show the pattern and launches more helper agents for the larger ones.
  \item[\textbf{33--40.}] It continues expanding coverage, loading \texttt{all-icetea-ai-research}, another general channel, \texttt{bfcl-prompt-v-01}, and \texttt{cognitive-
  paper-code-base}, and again launches helper agents.
  \item[\textbf{41--48.}] It checks the remaining channels, including \texttt{hypothesis-literature-agent}, \texttt{tool-bench-langvar-dev}, \texttt{tool-bench-langvar}, and
  a few low-signal channels; it concludes that social/random channels are irrelevant.
  \item[\textbf{49--59.}] Because the question mentions code review, it decides to inspect GitHub evidence as well. It searches Slack for GitHub repos, then uses tool search
  to select thread and GitHub PR/review tools.
  \item[\textbf{60--76.}] It searches GitHub repositories mentioned in Slack and lists pull requests across several repos, including \texttt{iceteaResearchChannel},
  \texttt{neurips\_operator}, \texttt{bfcl-prompt-v-01}, \texttt{research-ideagen}, and \texttt{misalign-capabilities}. This does not surface a decisive peer-dismissal pattern.
  \item[\textbf{77--80.}] It returns to \texttt{bfcl-more-tools}, noting that this channel had previously produced errors. It retries the channel history, but again
  remarks that access is unreliable.
  \item[\textbf{81--83.}] It runs more targeted BFCL/result-oriented searches such as \texttt{doubt question methodology flawed}, \texttt{doesn't make sense skeptical
  disagree numbers}, and \texttt{Marcus results findings}. These searches still do not produce a clear answer.
  \item[\textbf{84--91.}] It re-examines thread-level interactions in the cognitive and paper-operator projects. It initially considers whether Ray reviewing Harper's work
  might fit, but then rejects that hypothesis after seeing the reviews are consistently positive.
  \item[\textbf{92--100.}] It tries more BFCL-specific and generic criticism searches, including \texttt{Ryan results in:bfcl-more-tools}, \texttt{Brandon results wrong
  thousand}, \texttt{not convinced}, \texttt{doubt suspicious misleading}, and \texttt{every time always keeps}. These again fail to identify the intended pair.
  \item[\textbf{101--107.}] It pivots back to the cognitive channels and uses Bash analysis over saved histories. It first concludes that \texttt{cognitive-foundations}
  looks collaborative, then starts tracking who finds bugs in whose code inside \texttt{cognitive-paper-code-base}. At this point it explicitly notices that Chris repeatedly finds
  bugs that Alex ends up fixing.
  \item[\textbf{108--109.}] It checks \texttt{bfcl-prompt-v-01} more carefully and explicitly concludes that Brandon and Ryan are collaborating well, with Brandon praising Ryan's
  work. It therefore rules out that channel as the answer.
  \item[\textbf{110--112.}] It finally gets another look at \texttt{bfcl-more-tools} and concludes that Brandon is praising Marcus there and that the BFCL interaction
  pattern is collaborative rather than dismissive. This is the critical moment where it effectively discards the correct social cluster.
  \item[\textbf{113--120.}] It rechecks \texttt{cognitive-foundations}, \texttt{tool-bench-langvar-dev}, \texttt{misalign-capabilities}, and \texttt{research-
  ideagen}. It repeatedly says these channels look collaborative or are dominated by project-lead-to-RA dynamics.
  \item[\textbf{121--124.}] It returns to \texttt{cognitive-paper-code-base} one more time and manually enumerates a sequence of Chris's pointed bug reports against Alex's
  code. Even here, it still acknowledges that the discussion looks like normal collaborative review in isolation.
  \item[\textbf{125--127.}] It runs three final generic searches: \texttt{again same mistake}, \texttt{not buying questionable approach}, and \texttt{don't think this is
  right}. These do not produce a stronger alternative.
  \item[\textbf{Step 128.}] The agent explicitly says that it has searched all channels and still has not found a clear peer-to-peer dismissal pattern, but because the question
  says someone flagged a concern, ``it must exist somewhere.'' It then commits to the Chris$\rightarrow$Alex pattern as the best locally coherent explanation.
  \item[\textbf{Step 129.}] It returns the final answer that Chris is dismissing Alex.
  \end{enumerate}
  }

  \textbf{Failure analysis.}
  This is a clear example of \emph{premature local closure instead of recurring-theme analysis}. The agent does not fail because it searched too little: it inspects nearly every
  substantive Slack channel, launches helper agents, checks thread structure, and even looks at GitHub pull requests. The failure happens later, when it must compare competing
  interpersonal patterns across candidates. Once the BFCL channels are noisy, partially inaccessible, or superficially collaborative, the agent stops aggregating weaker recurring
  signals there and instead locks onto a single, vivid local pattern in \texttt{cognitive-paper-code-base}: Chris repeatedly finding bugs in Alex's code with sharp language. That
  pattern is coherent enough to support a narrative, so the agent treats it as decisive. The intended benchmark signal, however, is subtler and distributed across BFCL
  interactions: the workspace modification explicitly adds repeated cases where Ryan questions Marcus's results after Marcus posts them, but not Brandon's. The agent therefore
  exhibits local closure: after broad search, it commits to the first channel-local story that seems to fit, rather than performing a global recurrence comparison over the
  relevant candidate pairs.
\section{Prompts}
\label{appendix:prompts}

We provide the prompt for LLM judge here: 

 System prompt:

  You are a helpful assistant that judges the final answer of a LLM to a given query.
  You will be given the query, the final answer, and the reference answer.
  You will need to judge if the final answer is correct according to the query and the reference answer.
  You will need to output your judgment in the format: JUDGMENT: <your judgment>.
  Your judgment should be based on the key information correctness, for example, if the query asks for names, only check the correctness of the names, not the exact match of the
  whole final answer. Also, for more open questions, you should only check the key information correctness! For example, if the query ask for things like ``what strategy did they
  use'', as long as the strategy is correct, it is fine, you don't need to check the exact match of the whole final answer.
  Examples of key information correctness:
  1. reference answer: ``gpt-4.1-2025-04-14'', model answer ``gpt-4'', INCORRECT.
  Please only output your judgment. Your judgment should be either ``YES'' for correct or ``NO'' for incorrect.

  User prompt:

  QUERY: \{query\}\\
  FINAL\_ANSWER: \{final\_answer\}\\
  REFERENCE\_ANSWER: \{reference\_answer\}\\
  Please judge if the FINAL\_ANSWER is correct according to the QUERY and the REFERENCE\_ANSWER.
  Warning again: Your judgment should be based on the key information correctness, for example, if the query asks for names, only check the correctness of the names, not the exact
   match of the whole final answer. Also, for more open questions, you should only check the key information correctness! For example, if the query ask for things like ``what
  strategy did they use'', as long as the strategy is correct, it is fine, you don't need to check the exact match of the whole final answer.
  Examples of key information correctness:
  1. reference answer: ``gpt-4.1-2025-04-14'', model answer ``gpt-4'', then model answer is INCORRECT.
  Your response:
\section{Agent Harnesses and Models}
\label{appendix:agents_models}

\subsection{Agent Harnesses}

We evaluate six command-line agents: Claude Code \citep{anthropic2025claudecode}, Codex CLI \citep{openai2025codex}, Gemini CLI \citep{google2025geminicli}, Antigravity CLI \citep{google2025antigravity}, Grok Build \citep{xai2026grokbuild}, Muse Code \citep{meta2026musecode}, and two open-source agents: OpenHands \citep{wang2025openhandsopenplatformai}, Terminus-2 \citep{merrill2026terminalbenchbenchmarkingagentshard}. Unless otherwise specified, the reasoning effort is set to medium. We set a time limit of 20 minutes for each query.

\subsection{Models}

We select the following frontier models: Gemini-4-Argon \citep{kavukcuoglu2026gemini4argon}, Claude-Fable-5.1 \citep{anthropic2026fable51}, Muse Wonder \citep{meta2026musewonder}, Claude-Opus-5.5 \citep{anthropic2026opus55}, GLM-5.3 \citep{zai2026glm53}, Grok 4.7 \citep{spacexsi2026grok47}, Muse-Spark-1.3 \citep{meta2026musespark13}, GPT-6.1-Terra \citep{openai2026gpt61terra}, GPT-6.1-Sol \citep{openai2026gpt61sol}, GPT-6-Astra \citep{openai2026gpt6astra}, GPT-6-Luna \citep{openai2026gpt6luna}, Qwen3.7-Max \citep{alibabacloud2026qwen37max}, Claude-Opus-4.7 \citep{anthropic2026claudeopus47}, Claude-Opus-4.6 \citep{anthropic2026opus46}, Claude-Sonnet-4.6 \citep{anthropic2026sonnet46}, Claude-Haiku-4.5 \citep{anthropic2025haiku45}, GPT-5.5 \citep{openai2026gpt55}, GPT-5.4 \citep{openai2026gpt54}, GPT-5.4-mini \citep{openai2026gpt54mininano}, Gemini-3.1-Pro-Preview \citep{googledeepmind2026geminipro}, Gemini-3-Flash-Preview \citep{googledeepmind2026geminiflash}, GLM-5.1 \citep{glm5team2026glm5vibecodingagentic}, Kimi K2.6 \citep{moonshot2026kimik26}, DeepSeek-V4-Pro \citep{deepseekai2026deepseekv4}. We access all models through their first-party APIs.

\end{document}